\documentclass[10pt,journal,compsoc]{IEEEtran}
\ifCLASSOPTIONcompsoc
  \usepackage[nocompress]{cite}
\else
  \usepackage{cite}
\fi
\usepackage{hyperref}
\usepackage{url}
\usepackage{microtype}
\usepackage{graphicx}
\usepackage{amsfonts}
\usepackage{amsmath}
\usepackage{amsthm}
\usepackage{booktabs}
\usepackage{algorithm}
\usepackage{algorithmic}
\usepackage{wrapfig}
\usepackage{color}
\usepackage{bm}
\usepackage{xr}

\def\myname{DH-Graph2Seq}

\ifCLASSINFOpdf
\else
\fi
\begin{document}
%
\title{Ask Questions with Double Hints: Visual Question Generation with Answer-awareness and Region-reference}
%
%
%
%

\author{Kai~Shen, Lingfei~Wu,~\IEEEmembership{Member,~IEEE,}
Siliang~Tang, 
Fangli~Xu, 
Bo~Long, 
Yueting~Zhuang, 
Jian~Pei,~\IEEEmembership{Fellow,~IEEE}
\IEEEcompsocitemizethanks{
\IEEEcompsocthanksitem Kai Shen, Siliang Tang and Yueting Zhuang are with Zhejiang University, Hang Zhou, China. Email: \{shenkai, siliang, yzhuang\}@zju.edu.cn. Corresponding author: Siliang Tang.\protect\\
\IEEEcompsocthanksitem Lingfei Wu and Fangli Xu are with Anytime AI. Email: \{teddy.lfwu, fanglixu55\}@gmail.com.\protect\\
\IEEEcompsocthanksitem Bo Long is with Meta AI. Email: bo.long@gmail.com.\protect\\
\IEEEcompsocthanksitem Jian Pei is with Duke University. Email: j.pei@duke.edu.\protect\\
}
}

%
%

\markboth{JOURNAL OF LATEX CLASS FILES, VOL. XXX, NO. XXX, SEPTEMBER 2021}%
{Shell \MakeLowercase{\textit{et al.}}: Bare Demo of IEEEtran.cls for Computer Society Journals}
%



\IEEEtitleabstractindextext{%

\begin{abstract}
The visual question generation~(VQG) task aims to generate human-like questions from an image and potentially other side information (e.g. answer type). Previous works on VQG fall in two aspects: i) They suffer from \textit{one image to many questions mapping} problem, which leads to the failure of generating referential and meaningful questions from an image. ii) They fail to \textit{model complex implicit relations} among the visual objects in an image and also overlook potential interactions between the side information and image. 
To address these limitations, we first propose a novel learning paradigm to generate visual questions with answer-awareness and region-reference. 
Concretely, we aim to ask the right visual questions with \emph{Double Hints - textual answers and visual regions of interests}, which could effectively mitigate the existing one-to-many mapping issue. Particularly, we develop a simple methodology to self-learn the visual hints without introducing any additional human annotations. Furthermore, to capture these sophisticated relationships, we propose a new double-hints guided Graph-to-Sequence learning framework, which first models them as a dynamic graph and learns the implicit topology end-to-end, and then utilizes a graph-to-sequence model to generate the questions with double hints. 
Experimental results demonstrate the priority of our proposed method.
\end{abstract}

}

\maketitle

\IEEEdisplaynontitleabstractindextext

%
\IEEEpeerreviewmaketitle

\IEEEraisesectionheading{\section{Introduction}\label{sec:intro}}

%
%
%
%

 

\IEEEPARstart{V}isual Question Generation (VQG) is an emerging task in both computer vision~(CV) and natural language processing~(NLP), which aims to generate human-like questions from an image and potentially other side information (e.g. answer type or answer itself). Recently, there has been a surge of interests in VQG because it is particularly useful for providing high-quality synthetic training data for visual question answering~(VQA)~\cite{li2018visual,10.1109/TPAMI.2017.2754246,7934440} and visual dialog system~\cite{Jain_2018_CVPR}. Conceptually, it is a challenging task because the generated questions are not only required to be consistent with the image content but also meaningful and answerable by humans.

Despite promising results that have been achieved, the \textit{one-to-many mapping} and \textit{complex implicit relation modelling} problems hinder the development of visual question generation.

The \textit{one-to-many mapping problem} occurs when many potential questions can be mapped to certain inputs, which betrays the supervised objective which is usually one certain question. This phenomenon leads to severe ambiguity preventing the model from producing the referential and meaningful questions from an image. 
Conceptually, the existing VQG methods can be generally categorized into three classes with respect to what hints are used for generating visual questions: 1) the whole image as the only context input~\cite{mora2016towards}; 2) the whole image and the desired answers~\cite{li2018visual}; 3) the whole image with the desired answer types~\cite{krishna2019information}. Since a picture is worth a thousand words, it can be potentially mapped to many different questions, leading to the generation of diverse non-informative questions with poor quality. Even with the answer type or desired answer information, the similar one-to-many mapping issue remains, partially because the answer hints are often very short or too broad.

\begin{figure}[tb]
  \centering
  \includegraphics[width=1\linewidth]{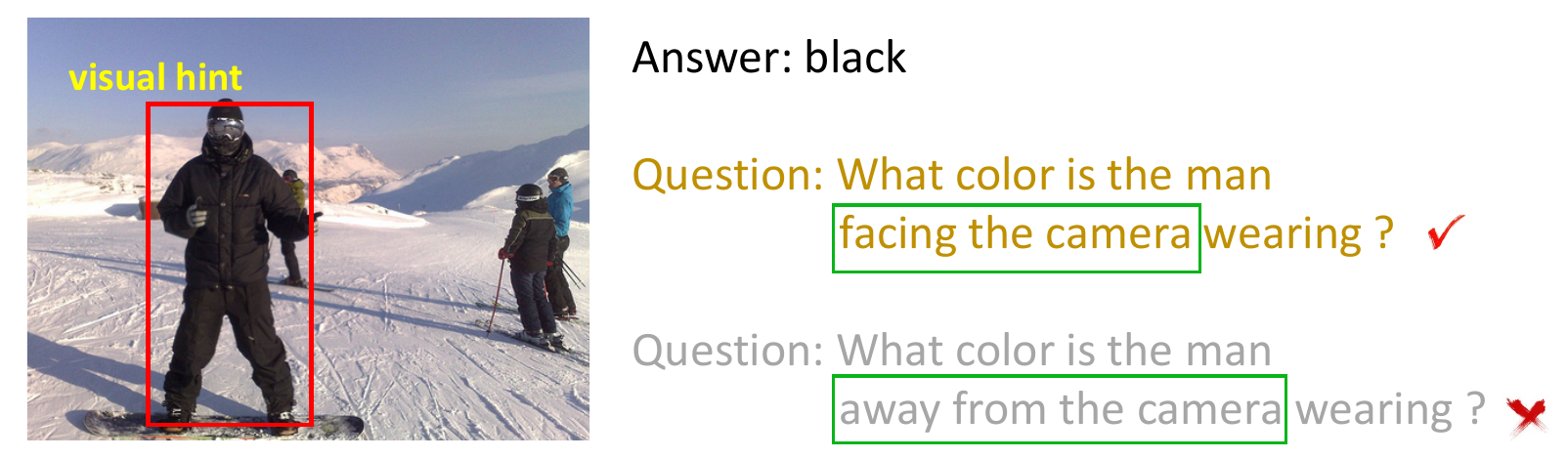}
  \caption{An example of our new setting for learning VQG with \emph{Double Hints: textual answers and visual regions of interests}.}
  \label{intro}
\end{figure}

As Figure \ref{intro} shows, one person is facing the camera while another person is away from the camera, but both of them wear black clothes. The answer here is \textit{black color} and we need to ask a right question given this answer. In this case, asking question about either of them could be acceptable when the answer information is the only hint. As a result, existing side information is often not informative enough for guiding question generation process, which causes the failure of generating referential and meaningful questions from an image. Therefore, the visual regions of interest hints are often crucial for helping reduce the ambiguities in order to ask a right question. Based on both the textual answer hints and visual regions of interest hints, we can significantly mitigate the existing one-to-many mapping issue of VQG.

\begin{figure}[tb]
  \centering
  \includegraphics[width=1\linewidth]{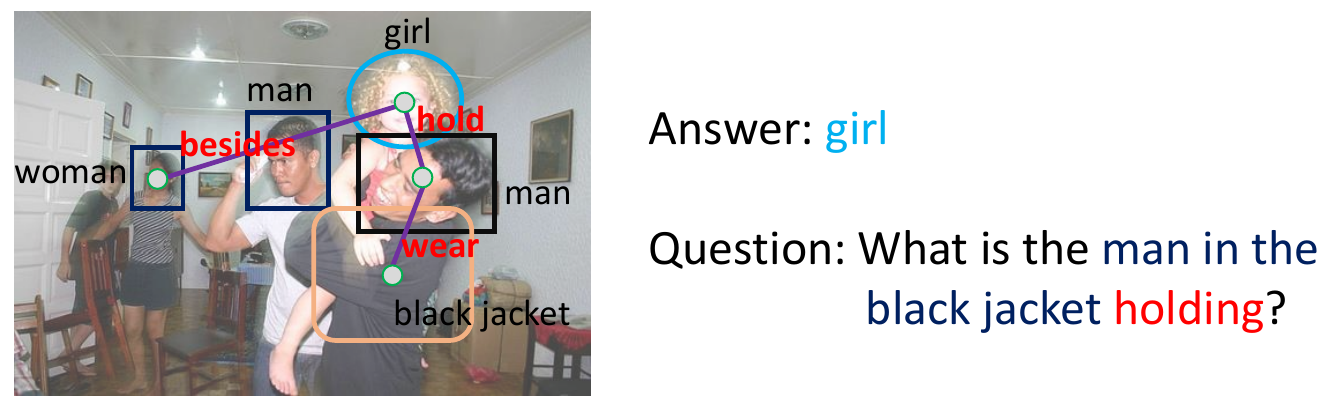}
  \caption{An example of the correlations among the visual objects.}
  \label{intro_relation_graph}
\end{figure}

The second severe issue is due to \textit{complex implicit relation modelling} problem in VQG. This is partially because that the existing VQG methods often ignore the rich correlations among the visual objects in an image and potential interactions between the side information and image~\cite{krishna2019information}.
Conceptually, the implicit relations among the visual objects (e.g., spatial, semantic) could be the key to generating meaningful and high-quality questions. For example, as Figure \ref{intro_relation_graph} shows, the girl has multiple relations (hold, besides) with the other men, which provide vital clues for generating high-quality questions. 
In addition, another important factor for producing informative and referential questions is how to make full use of the side information to align with the targeted image. In this example, 
given the target answer girl, the relation \textit{hold} between the man in the front and the target answer \textit{girl} greatly contributes to the question generation process.
In consequence, modeling such potential interactions between the side information and an image becomes a critical component for generating referential and meaningful questions. 

To address the first issue, we first propose a novel learning paradigm for the visual question generation task, which generates the visual questions with \emph{Double Hints - textural answer and visual regions of interests}. 
Thanks to the proposed double hints, the ambiguities are largely reduced since the questions can be clearly mapped to the specific answer and visual hints.
Concretely, we aim to utilize the referential visual regions of interest hints~(denoted as visual hints for simplicity) of the images and the textual answers~(denoted as answer hints) to faithfully guide question generation. As illustrated in Figure \ref{intro}, by giving an image with visual hints (the region enclosed by the orange rectangle) and answer hints (the answer), the model is able to  generate the right question with key entities that reflects the visual hints and is answerable to the answer hints. To this end, in order to learn these visual hints, we develop a multi-task auto-encoder to self-learn the visual hints and the unique attributes automatically without introducing any additional human annotations.

Furthermore, to address the second issue, to capture the rich interactions between double hints and the image, as well as the sophisticated relationships among the visual objects, we propose a new Double-Hints guided Graph-to-Sequence learning framework (\myname{}). The proposed model first models these interactions as a dynamic graph and learns the implicit topology end-to-end. Then it utilizes a Graph2Seq model to generate the questions with double hints. In addition, in the decoder side, we also design a visual-hint guided separate attention mechanism to attend image and object graph separately and overlook the non-visual-hints particularly.

In summary, we highlight our main contributions as follows:
\begin{itemize}
    \item We propose a novel learning paradigm to generate visual questions with \emph{Double Hints - textual answer and visual regions of interests}, which could effectively mitigate the one-to-many mapping issue. To the best of our knowledge, this is the first time both visual hints and answers hints are used for the VQG task.
    \item We explicitly cast the VQG task as a Graph-to-Sequence~(Graph2Seq) learning problem. We employ graph learning technique to learn the implicit graph topology to capture various rich interactions between and within an image, and then utilize a Graph2Seq model to guide question generation with double hints.
    \item Our extensive experiments on VQA2.0 and COCO-QA datasets demonstrate that our proposed model can significantly outperform existing state-of-the-art by a large margin. Further experiments show that our VQG can help VQA as a data augmentation method when the training data is limited.
\end{itemize}

\begin{figure*}[h]
  \centering
  \includegraphics[width=0.85\linewidth]{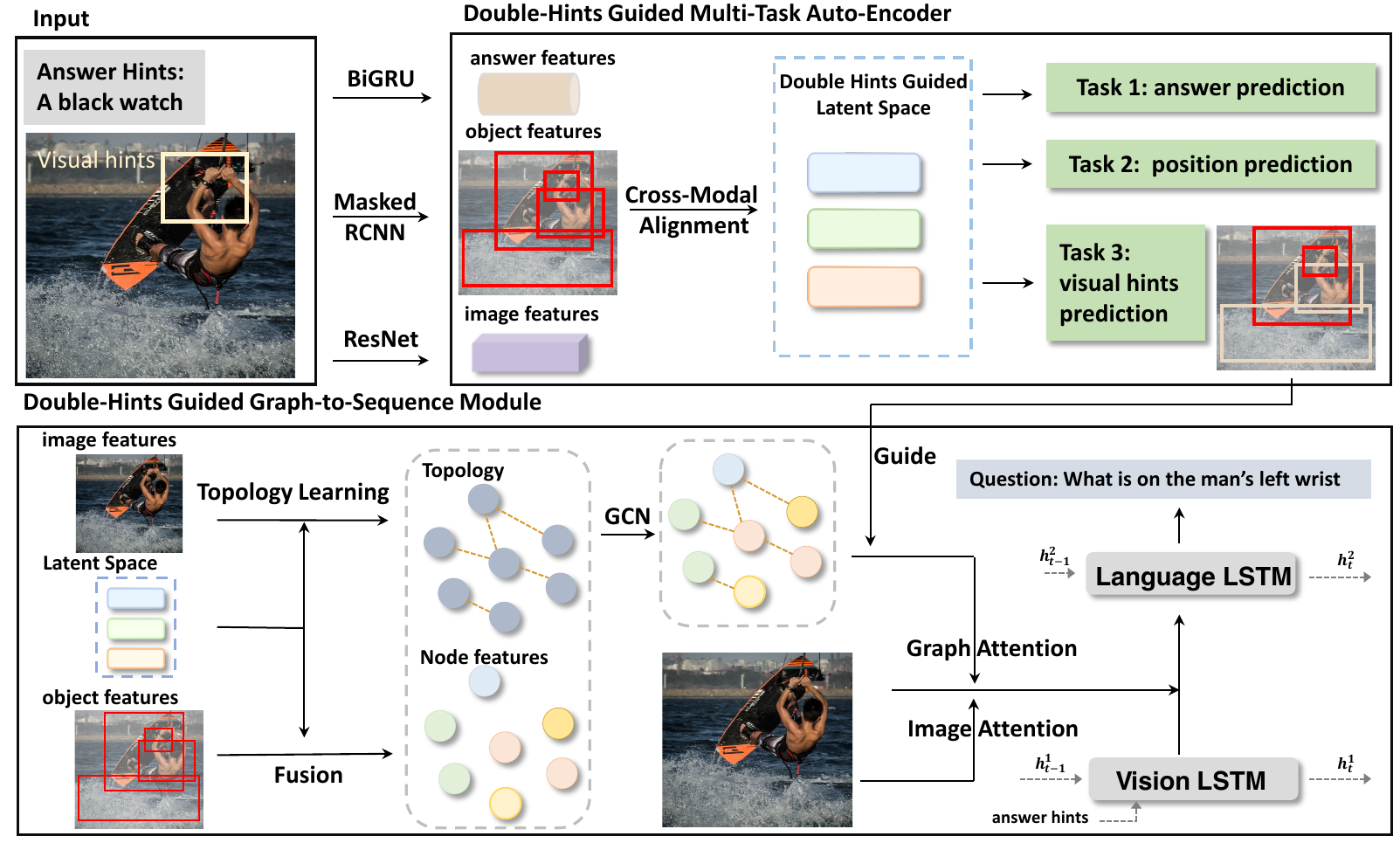}
  \caption{The overall framework of our proposed model with double hints to guide VQG.}
  \label{fig:framework}
\end{figure*}

\section{Related Work}
\subsection{Visual Question Generation}

Visual question generation is an emerging task in the visual-language domain~\cite{mora2016towards,mostafazadeh2016generating,9126124,8847465,8528867}.
Mora et al.~\cite{mora2016towards} firstly made an attempt to generate question-answer pairs based on an image to support the visual question answering (VQA) task. They simply employed the VGG network as the encoder and use the long short-term memory
networks (LSTM) to produce both the questions and answers outputs. Mostafazadeh et al.~\cite{mostafazadeh2016generating} formally defined the VQG task. Different from straightforward questions, they focused on questions that can potentially engage a human in starting a conversation. In order to generate more diverse questions, Jain et al.\cite{jain2017creativity} combined the advantage of variational autoencoders with LSTMs to entrust the model with creativity. Compared with conventional methods, it can generate a large set of varying questions given a single input image.
Zhang et al.\cite{zhang2016automatic} proposed to generate questions that can be visually grounded in the given images. They firstly generated the textural captions from the images and then predict the specific question types. Secondly, the final question was generated with the guidance of both the question type and the caption. These works simply mapped the visual images into the textural questions, which lead to imprecise and generic questions. 

Liu et.al.\cite{liu2018ivqa} viewed this task the inverse of VQA, which regarded the target answers as the guidance to generate high-quality questions. They posed the question generation as a multi-modal dynamic inference process to help improve the questions' quality. What's more, the dual learning mechanism is applied in the VQG task. In \cite{li2018visual}, the VQA and VQG are considered as dual tasks and are trained jointly in a unified end-to-end framework. Furthermore, the parameter sharing and regular techniques are proposed as constraints to leverage the question-answer dependency. Shah et al.\cite{shah2019cycle} employ the VQG to boost the VQA models' robustness. Different from conventional methods, Krishna et al.\cite{krishna2019information} introduce the fine-grained answer type as the guidance to the variational method, which generates goal-driven questions. Xu et al.\cite{xu2020radial} adapt the graph method to generate meaningful questions with the target of answers. Different from these works, we will generate the questions under the guidance of both visual regions of interest and textual answers, which can generate more referential and answerable questions.

\subsection{Graph-to-Sequence Learning}
Graph neural networks~(GNN) ~\cite{kipf2016semi,gilmer2017neural} has drawn a significant amount of attention in recent years. In the NLP domain, graph-to-sequence learning is to generate sequential results from graph-structured data, which maps the structural data to sequence output~\cite{xu2018graph2seq,chen2019reinforcement,gao2019dyngraph2seq}. When coming to the non-graph structured data just like the regions of an image, researchers explore some methods to construct objects'~\cite{norcliffe2018learning} or words'~\cite{chen2019graphflow,chen2019reinforcement} topology. 

Unlike these previous methods, we propose a novel cross-modal graph2seq model which models the relations upon the visual and textural hints. With the guidance of these two modalities, the model is able to learn the appropriate embeddings which are crucial to generate high-quality questions.

\section{Method}
We first introduce our problem formulation and then the proposed learning paradigm that generates visual questions with double hints. Next, we discuss each key component of our overall framework of proposed model in Figure \ref{fig:framework}.

\subsection{Problem Formulation}
The goal of the visual question generation task is to generate human-like and answerable natural language questions based on the given images and potentially other side information, such as textural answer and answer type. The generated questions need to be consistent with the given images and the hints semantically. In this part, we will first discuss the traditional answer-based hints and then present our new learning paradigm.

We assume that the raw image is $I$, and the target answer~(answer hint) is a collection of word tokens represented by $A = \{a_1, a_2, ..., a_m\}$, where $m$ denotes the total number of the answer words. The task of traditional answer-guided VQG is to generate the natural language question consisting of a word tokens sequence  $Q = \{q_1, q_2, ..., q_n\}$ which maximizes the conditional likelihood:
\begin{equation}
    \hat{Q} = \text{argmax}_{Q}P(Q|I, A),
\end{equation}
 where $n$ is the total number of question words. 
To address the existing one-to-many mapping issue, we introduce a new setting that focuses not only on the answer-awareness but also on region-reference. 
Specifically, we cast it as a joint optimization problem over visual hints finding tasks and double-hints guided graph2seq learning task. The visual hint is a collection of visual object regions of interest in the image which are served as direct visual clues for visual question generation. 
We denote it as $\mathcal{V}=\{v_1, v_2, ..., v_N\}$, where $v_i$ is the bounding box in the image and $N$ is the number of visual hints. Under this setting, the likelihood can be cast as:
\begin{equation}
    P(Q|I, A) = P(Q|\mathcal{V}, I, A) P(\mathcal{V}|I, A).
\end{equation}
In particular, the visual hints will be generated in the data pre-processing phase~(see Sec. \ref{sec:dataset}) and the model can automatically learn them without any human annotations. Thus there is no additional visual hints annotation during inference.

\subsection{Image and Answer Encoder}
\label{sec:image_and_answer_encoder}
\textbf{Image Encoder.}
Given a raw image $I$, we first extract the image feature by a pretrained convolutional neural network~\cite{he2016deep} given by $\mathbf{I} \in \mathbb{R}^{c \times F_v}$ where $c$ is the channel number and $F_v$ is the visual feature dimension. Secondly, we assume that the image is represented by a collection of object regions denoted by $\mathcal{R} = \{v_1, v_2, .., v_T\}$. 
The object regions are associated with visual feature $\mathbf{V} = [\mathbf{v}_1, \mathbf{v}_2, ..., \mathbf{v}_T]$, where $\mathbf{v}_i \in \mathbb{R}^{F_v}$ and $F_v$ is the feature dimension. In addition, each object $v_i$ has addition bounding box position vector $p_i = (x_0, y_0, x_1, y_1)$, category attribute~(eg. it belongs to animal class) $c_i$. $(x_0, y_0), (x_1, y_1)$ are the normalized up-left and bottom-right coordinates. It is worth noting that the visual hints $\mathcal{V}$ is the subset of visual regions collection $\mathcal{R}$ ($\mathcal{V} \subseteq \mathcal{R}$).

\noindent \textbf{Answer Encoder.}
As for the textural answers $A = \{a_1, a_2, ..., a_m\}$, we first employ the pretrained GloVe weights to initialize the embeddings which are represented by $\mathbf{A}_{initial} = [\mathbf{a}_1, \mathbf{a}_2, ..., \mathbf{a}_m]$. Then we adopt a GRU module to learn the word semantic features given by:
\begin{equation}
    \mathbf{A} = \text{GRU}(\mathbf{A}_{initial}),
\end{equation}
where $\mathbf{A} \in \mathbb{R}^{m\times F_a}$ and $F_a$ denotes the answer vector dimension. 

\subsection{Double-Hints Guided Multi-Task Auto-Encoder}
\label{sec:auto-encoder}
How to effectively find the visual hints and how to combine the visual features with double hints are particularly important for addressing the first issue. In this section, we first introduce a cross-modal alignment that aligns the cross-modal visual objects and textural answers into the latent space. Then we propose a multi-task decoder that can effectively predict the visual hints and the answer attributes from the learned latent space.

\subsubsection{Cross-modal Alignments}
\label{sec:align}
How to infer which regions of interest are suitable for asking questions from the answer clues and visual object features is a critical task. 
Intuitively, exploiting the rich fine-grained interactions between visual clues and textual answer hints is beneficial to find out the vital visual hints. To this end, we explicitly model the global correlations between them by the cross-modal alignment technique in the embedding space.

Firstly, we observe that the position and category attributes of objects are indispensable during fine-grained object relation modeling. Thus, for object $v_i$, we incorporate them by projecting the position vector $p_i$ and category attribute $c_i$ into two embedding spaces denoted as $\mathbf{p}_i \in \mathbb{R}^{d}$ and $\mathbf{c}_i \in \mathbb{R}^{d}$. Then we concatenate them with visual feature $\mathbf{v}_i$ given by:
\begin{equation}
    \mathbf{v}_i = \text{F}(\text{CONCAT}(\mathbf{v}_i,\mathbf{p}_i,\mathbf{c}_i)),
\end{equation}
where $\text{F}(\cdot): \mathbb{R}^{2\times d + F_v} \rightarrow \mathbb{R}^{F_v}$ is the linear projection with ReLU activation function and $\text{CONCAT}(\cdot, ..., \cdot)$ denotes the vector concatenation operation.
We slightly overload the notation $\mathbf{v}_{i}$ here for simplicity in the subsequent sections.

For each object $v_i$ ($\mathbf{v}_i$) and answer word $a_j$ ($\mathbf{a}_j$), we will calculate the alignment score $\mathbf{S}_{ij}$ as follows:
\begin{equation}
    \mathbf{S}_{ij} = \sigma(\mathbf{v}_i \mathbf{W}_r + \mathbf{a}_j \mathbf{W}_a) \mathbf{W},
\end{equation}
where $\sigma(\cdot)$ denotes the Tanh function, $\mathbf{W}_r \in \mathbb{R}^{ F_v \times F_{align}}$, $\mathbf{W}_a \in \mathbb{R}^{F_{a} \times F_{align}}$ and $\mathbf{W} \in \mathbb{R}^ {F_{align} \times 1}$, where $F_{align}$ is the hidden dimension size. The alignment score matrix $\mathbf{S} \in \mathbb{R}^{N \times m}$ represents the correlations' weight between the objects and the answer words. Following this correlation matrix, we can aggregate the answer words ($\{a_1, a_2, ..., a_m\}$) by visual object ($v_j$) given by:
\begin{equation}
\begin{split}
    \alpha_{ij} &= \frac{exp(\mathbf{S}_{ij})}{\sum_{k=0}^{k=m}{exp(\mathbf{S}_{ik})}}, \\
    \mathbf{a}'_i &= \sum^{m}_{j = 1} \alpha_{ij} \mathbf{a}_j.
\end{split}
\end{equation}
By this operation, the answer words having strong correlations with the specific objects are assigned with high alignment scores. Then we fuse the visual features with the aggregated textural answer features as follows:
\begin{equation}
    \mathbf{\overline{v}}_i = \text{Align}(\mathbf{v}_i, \mathbf{a}'_i) = \text{F} ( \text{CONCAT}(\mathbf{v}_i,\mathbf{a}'_i)),
    \label{eq:v_overline_latent}
\end{equation}
where $\mathbf{\overline{v}}_i$ is the aligned object representation, $\text{CONCAT}(\cdot, \cdot)$ is the vector concatenation operation and $\text{F}(\cdot): \mathbb{R}^{F_v + F_a} \rightarrow \mathbb{R}^{F_v}$ is the linear projection with ReLU nonlinearity.

\subsubsection{Multi-Task Decoder}
After aligning the visual features with the answers, we introduce the multi-task decoder which can infer the visual hints as well as  the unique attributes~(i.e the objects' position and category attributes and the target answers). Our hope is that the latent embedding could absorb the double hints information while retaining the robust features like objects' position clues.

Firstly, we apply MLP with ReLU activation to project the aligned visual objects' features to a $F_h$ dimension latent space denoted as $\mathbf{\overline{V}} = [\mathbf{\overline{v}}_1, \mathbf{\overline{v}}_2, ..., \mathbf{\overline{v}}_T], \mathbf{\overline{V}} \in \mathbb{R}^{T \times F_h}$. 
Empirically, we find it is beneficial to fuse the coarse-grained image features (i.e., $\mathbf{I} \in \mathbb{R}^{c \times F_v}$ in Sec. \ref{sec:image_and_answer_encoder}) with the fine-grained object features (i.e., $\mathbf{\overline{V}} \in \mathbb{R}^{T \times F_h}$). For each object feature $\mathbf{\overline{v}_i}$, we capture the top-down signals~\cite{anderson2018bottom} by attending to the image features as follows:
\begin{equation}
    \centering
    \begin{split}
        \mathbf{M}_{i, j} &= \sigma(\mathbf{\overline{v}}_i \mathbf{W}_1 + \mathbf{I}(j) \mathbf{W}_2 ) \mathbf{W}_3, \\
        \alpha_{i, j} &= \frac{exp(\mathbf{M}_{ij})}{\sum_{k=1}^{c}{exp(\mathbf{M}_{i,k})}}, \\
        \mathbf{o}_i &= \sum^{c}_{j=1} \alpha_{i, j} \mathbf{I}(j) \mathbf{W},
    \end{split}
\end{equation}
where $\mathbf{I}(j) \in \mathbb{R}^{F_v}$ denotes the $j$-th channel's feature vector, $\sigma(\cdot)$ is the Tanh function, and $\mathbf{W}_1 \in \mathbb{R}^{F_h \times F_{hid}}$, $\mathbf{W}_2 \in \mathbb{R}^{F_v \times F_{hid}}$, $\mathbf{W}_3 \in \mathbb{R}^{F_{hid}} \times 1$, $\mathbf{W} \in \mathbb{R}^{F_v \times F_h}$ are learnable parameters. Then we use residual connection to combine the aligned feature $\mathbf{\overline{v}}_i$ (Eq. \ref{eq:v_overline_latent}) and top-down feature $\mathbf{o}_i$ given by:
\begin{equation}
    \mathbf{\widetilde{v}}_i = \mathbf{\overline{v}}_i + \mathbf{o}_{i}.
    \label{eq:v_widetilde}
\end{equation}
In what follows, we will discuss how the multi-task decoder could generate the visual hints while keeping the unique attributes in the latent embedding.

\paragraph{Visual hint prediction} We first predict the probability of the object $v_i \in \mathcal{R}$ being a visual hint given by:
\begin{equation}
    \begin{split}
        s_{i} &=  \text{MLP}(\mathbf{\widetilde{v}})_i, \\
        P(i) &= \frac{exp(s_i)}{ \sum^{T}_{j=1} exp(s_j)},
    \end{split}
\end{equation}
where $\text{MLP}(\cdot): \mathbb{R}^{F_h} \rightarrow \mathbb{R}$ is the multi-layer projection with ReLU nonlinearity and $\widetilde{v}_i$ is the visual object features mentioned in Eq. \ref{eq:v_widetilde}.
In practice, the number of visual-hint objects is much smaller than non-visual-hint objects, thus we combine balanced-cross-entropy loss and focal loss~\cite{lin2017focal} together as follows:
\begin{equation}
\begin{split}
    L_{vh} =  - \frac{\eta}{N_{pos}}\sum_{i} \hat{y}_i P( i)^{\lambda} log(P(i)) \\
    - \frac{\eta}{N_{neg}} \sum_{i}( 1- \hat{y}_i) (1 - P(i))^{\lambda} log(1 - P(i)),
\end{split}
\end{equation}
where $\hat{y}_i$ is 1 iff $v_i \in \mathcal{V}$ (0 otherwise), $N_{neg}$ denotes the number of non-visual-hints and $N_{pos}$ is the number of visual hints.

\paragraph{Object position prediction} For high-quality question generation, the relative spatial relations among objects are important clues. To keep these essential features in the latent space, we employ the task by predicting the coordinates of the objects. Just like the previous visual hints prediction, we apply the feed-forward layer on $\mathbf{\overline{v}_i}$ and predict the absolute normalized position coordinates $p_i'$. The loss $L_{pos}$ is the mean square loss function of $p_i'$ and ground-truth $p_i$.

\paragraph{Target answer prediction} Since the answer hint is important in guiding how to ask high-quality questions, we ensure that the latent embedding indeed retains this information via regenerating the answer. Technically, we formulate it as a classification problem. We denote the answer is $t_i \in \{t_0, t_1, ..., t_c\}$, where $i$ is the $i$-th sample, and $c$ is the amount of all answers in the training dataset. Firstly, we apply a dilated CNN with a max-pooling layer to get the representation as follows:
\begin{equation}
    \begin{split}
        \mathbf{w} = \text{MaxPool}( \text{ReLU}(\text{DilatedCNN}(\mathbf{\widetilde{V}}))).
    \end{split}
\end{equation}
Then we apply the feed-forward layer on the pooled vector $\mathbf{w}$ and softmax function to calculate the probability. Finally, we adopt the cross-entropy loss to calculate the answer prediction loss denoted as $L_{ans}$.

\subsection{Double-Hints Guided Graph Construction and Embedding Learning}
\label{sec:graph-embedding}
To capture the complex correlations among visual objects in the image, we regard objects as nodes in the object graph $\mathcal{G}$ and adopt the paradigm of graph learning. Typically, all GNN algorithms are operated directly on graph-structured data and then compute the corresponding graph node embeddings. However, there is no prior graph-structured data in VQG since the relations among different objects are not explicitly given. Therefore, we will first discuss how to construct an object graph $\mathcal{G}$'s topology with the guidance of double hints and then employ a GNN model to encode them.

The topology of a graph represents the relations of the nodes. In our setting, the graph edges are weighted and learnable, since $\mathcal{G}$ is essentially a dynamic graph. We take the aligned objected embedding $\mathbf{\overline{v}}_i$ of multi-task auto-encoder (in Eq. \ref{eq:v_overline_latent}) to exploit this graph topology since the objects' visual features have incorporated double hints information in the auto-encoder.

Inspired by~\cite{chen2019deep}, a good similarity metric function has been proved be learnable and expressively powerful for learning graph structure. As a result, we first calculate the dense similarity matrix $\mathbf{S}$ by multi-head weighted cosine similarity function as follows:
\begin{equation}
    \begin{split}
        \mathbf{s}_{ij}^p &= cos(\mathbf{w}_p \odot \mathbf{\overline{v}}_{i}, \mathbf{w}_p \odot \mathbf{\overline{v}}_{j}), \\
    \mathbf{S}_{i,j} &= \frac{\sum_{p=1}^{K}{\mathbf{s}_{i,j}^p}}{K},
    \end{split}
\end{equation}
where $\mathbf{w}_p \in \mathbb{R}^{F_h}$ is the learnable weight, $\odot$ is the Hadamard product, and $\mathbf{s}_{i,j}^p$ denotes the similarity between node $i$ and $j$ of head $p$. The model learns to highlight specific dimensions of the latent embedding space. $K$ is the number of heads. The final similarity results is computed as the mean of the similarity scores from different subspaces. Because the learned graph similarity matrix $\mathbf{S}$ is dense and ranges between [-1, 1], we adopt $\epsilon-$sparsing to make it non-negative and sparse. Specifically, we mask off $\mathbf{S}_{i,j}$~(i.e., set to zero) if it is smaller than the threshold $\epsilon$, which leads to a final sparse adjacency matrix $\mathbf{A}$.

It is desirable to further align the objects' features with the learned double hints embeddings. 
Therefore, the final node features will be represented as follows:
\begin{equation}
\begin{split}
    \mathbf{X} &= \text{F}(\mathbf{V}; \mathbf{A}) = \text{CONCAT}(\mathbf{V}, \text{Latent}(\mathbf{V}, \mathbf{A}))  \\
    &= \text{CONCAT}(\mathbf{V}, \mathbf{\widetilde{V}})
    \label{eq:X}
\end{split}
\end{equation}
where $\text{Latent}(\cdot,\cdot)$ is the latent space representation generated from the auto-encoder described in Sec. \ref{sec:auto-encoder}. And $\text{CONCAT}(\cdot,\cdot)$ is the node-wise concatenation. 

Next, we apply a multi-layer graph convolution network~(GCN) with residual connection~\cite{he2016deep} to effectively learn the node embedding for each object from the constructed object graph. 
The basic layer is shown as follows:
\begin{equation}
    \mathbf{X}^{out} =\frac{\sigma(\mathbf{D}^{-1/2}\mathbf{\hat{A}}\mathbf{D}^{-1/2}\mathbf{X}^{in}\mathbf{W}) + \mathbf{X}^{in} }{\sqrt{2}} 
\end{equation}
where $\mathbf{X}^{in}$ is the node feature~($\mathbf{\overline{V}}$ in Eq.7), $\mathbf{\hat{A}}$ is the adjacency matrix~($\mathbf{A} + \mathbf{I}$), $\mathbf{D}$ is the degree matrix of $\mathbf{\hat{A}}$, $\mathbf{W}$ is the trainable weights and $\sigma(\cdot)$ is the ReLU activation function.
Then we stack $k$ layers of classic spectral GCN~\cite{kipf2016semi} together with residual architecture to aggregate the fine-grained object features. 

\subsection{Double-Hints Guided Question Generation}
In this section, we employ the attention-based hierarchical sequence decoder from~\cite{lu2018neural} for the double-hints guided question generation step with two alternative implementations: 1) LSTM based decoder and 2) transformer based decoder. We will discuss them respectively.
\subsubsection{LSTM based Question Generation Decoder}
\label{sec:ques-gen}
This module consists of two LSTMs: 1) vision LSTM and 2) language LSTM. The vision LSTM is used to encode the visual features and the language LSTM is used to generate words. Note that the starting states are initialized by answer features. Between the two LSTMs, the model attends on the image inputs and graph inputs separately guided by visual hints. We refer to this procedure as visual-hint-guided separate attention.

\noindent \textbf{Vision LSTM.} Technically, at time step $t$, we first adopt the vision LSTM to fuse the previous step's hidden state $\mathbf{h}^{t-1}_1$ with the global image features $\mathbf{I}$ and the word embedding $\mathbf{q}_t$ to create a current step's hidden state $\mathbf{h}^t_1$ as follows:
\begin{equation}
    \mathbf{h}^t_{1} = \text{LSTM}(\overline{\mathbf{I}}, \mathbf{q}_{t};\mathbf{h}^{t-1}_1),
\end{equation}
where $\mathbf{\overline{I}} \in \mathbb{R}^{F_v}$ is the mean pooling result of image feature $\mathbf{I}$.

\noindent \textbf{Visual-hint-guided Separate Attention.} The visual-hint-guided separate attention module then attends to the coarse-grained image feature and fine-grained graph features, respectively. For image attention, we apply the classic attention mechanism on the image feature given by:
\begin{equation}
    \mathbf{h}_{image} = \text{Attention}(\mathbf{I}, \mathbf{h}_{1}^{t}),
\end{equation}
where $\text{Attention}(\cdot,\cdot)$ is the classic attention mechanism~\cite{bahdanau2014neural}. For graph attention, we ignore the nodes which are predicted to be non-visual-hint. Therefore, we can define our graph attention mechanism as follows:
\begin{equation}
    \begin{split}
        \mathbf{X}_{vh} &= \text{VisualHintMask}(\mathbf{X}) \\
        \mathbf{h}_{graph} &= \text{Attention}(\mathbf{X}_{vh}, \mathbf{h}_{1}^{t})
    \end{split}
\end{equation}
where $\text{VisualHintMask}(\cdot)$ is to mask off the non-visual-hint objects, $\mathbf{X}$ is the graph node embedding and $\text{Attention}(\cdot,\cdot)$ is also the attention function. 

\noindent \textbf{Language LSTM.} Finally, the language LSTM will absorb the image attention features $\mathbf{h}_{image}$ and graph attention result $\mathbf{h}_{graph}$ as follows:
\begin{equation}
    \mathbf{h}^t_{2} =  \text{LSTM}(\text{CONCAT}(\mathbf{h}_{image}, \mathbf{h}_{graph}, \mathbf{h}^t_1); \mathbf{h}^{t-1}_2),
\end{equation}
where the $\mathbf{h}_{image}, \mathbf{h}_{graph}$ and the vision LSTM's hidden state $\mathbf{h}^t_1$ are concatenated, $\mathbf{h}^{t-1}_2$ is the language LSTM's previous step hidden state and $\mathbf{h}^{t}_2$ is the generated hidden state. We project the $\mathbf{h}^t_{2}$ to the vocabulary space to predict the word. We train it with the cross-entropy loss $L_{ques}$.

\subsubsection{Transformer based Question Generation Decoder}
Another widely-adopted architecture for language generation is the transformer. Similar to the LSTM-based architecture, we apply visual-hint-guided separate attention by the multi-head attention mechanism. The transformer decoder consists of $n$ basic layers. 

Formally, we define the input word embeddings as $\mathbf{Q} = \{\mathbf{q}_1, \mathbf{q}_2, ..., \mathbf{q}_n\} \in \mathbb{R}^{n \times d}$, where $n$ is the number of answer words. 
In order to fuse the answer hints into the decoder, we additionally add the the mean pooling of answer features (denoted as $\mathbf{\overline{a}}$)
to the front of the $\mathbf{Q}$ as $\mathbf{Q} = \{\mathbf{\overline{a}}, \mathbf{q}_1, \mathbf{q}_2, ..., \mathbf{q}_n\}$ (we overwrite $\mathbf{Q}$ in this section to make it clear to illustrate).
The framework of the basic decoder layer is shown as follows:
\begin{equation}
    \begin{split}
        \mathbf{Q} &= \text{Norm}(\text{MSA}(\mathbf{Q}, \mathbf{Q}, \mathbf{Q}) + \mathbf{Q}), \\
        \mathbf{Q} &= \text{Norm}(\text{VHSA}(\mathbf{Q}, \mathbf{I}, \mathbf{X}_{vh}) + \mathbf{Q}), \\
        \mathbf{Q} &= \text{Norm}(\text{FeedForward}(\mathbf{Q})),
    \end{split}
\end{equation}
where $\mathbf{Q}$ is the word embedding matrix, $\mathbf{I}$ is the image feature, $\mathbf{X}_{vh}$ is the visual hints feature (the same as Sec. \ref{sec:ques-gen}), $\text{MSA}(\cdot,\cdot, \cdot)$ denotes the multi-head self-attention, $\text{VHSA}(\cdot,\cdot, \cdot)$ denotes the visual-hint-guided separate attention and $\text{FeedForward}(\cdot)$ denotes the feedforward layer. In detail, the $\text{VHSA}(\cdot,\cdot, \cdot)$ which denotes the visual-hint-guided separate attention module can be formulated as follows:
\begin{equation}
    \begin{split}
        \mathbf{Q}_{img} &= \text{Norm}(\text{MHA}(\mathbf{Q}, \mathbf{I}, \mathbf{I}) + \mathbf{Q}), \\
        \mathbf{Q}_{graph} &= \text{Norm}(\text{MHA}(\mathbf{Q}, \mathbf{X}_{vh}, \mathbf{X}_{vh}) + \mathbf{Q}), \\
        \mathbf{Q}_{all} &= \mathbf{Q} + \mathbf{Q}_{graph} + \mathbf{Q}_{img}, \\
        \mathbf{Q}_{out} &= \text{Norm}(\text{MSA}(\mathbf{Q}_{all}, \mathbf{Q}_{all}, \mathbf{Q}_{all}) + \mathbf{Q}_{all}),
    \end{split}
\end{equation}
where the $\text{MHA}(\cdot, \cdot, \cdot)$ denotes the multi-head attention. 

After decoding, we simply drop the first row of $\mathbf{Q}_{out}$ which is corresponding to the answer hints, and use the rest to calculate the cross-entropy loss $L_{ques}$.

The final overall loss is thus the combination of  all the previous key components,
\begin{equation}
    L = L_{ques} + \alpha L_{vh} + \beta L_{pos} + \gamma L_{ans}.
\end{equation}

\begin{table*}[t]
\centering
\caption{Results on VQA2.0 and COCO-QA val set. Ours-LSTM denotes the \myname{} with LSTM decoder. Ours-transformer denotes the \myname{} with transformer decoder. All accuracies are in \%. }
\begin{tabular}{l|ccccc|ccccc}
\toprule  
Dataset & \multicolumn{5}{|c}{VQA 2.0} &  \multicolumn{5}{|c}{COCO-QA}  \\
\midrule
Method & BLEU@4& CIDEr&METEOR & ROUGE &Q-BLEU@1 & BLEU@4& CIDEr&METEOR & ROUGE &Q-BLEU@1\\
\midrule  
I2Q & 9.02 & 63.21 & 13.89 & 35.33 & 26.32 &13.53& 95.90& 12.61& 36.23 & 31.37 \\ 
IT2Q & 18.41 &134.88 & 19.90 & 45.71 & 40.27 & 17.80& 128.64& 16.17& 43.22& 38.88\\
IMVQG &19.72 &149.28 & 20.43 & 46.76  & 40.40 & 18.43& 127.18& 17.21 & 44.07& 39.22\\
Dual&19.90&151.60 & 20.60 & 47.00  & 41.90 &18.80& 131.10& 17.73& 44.19& 39.92 \\
Radial  & 20.70& 161.90& 21.40 & 48.10& 43.50 & 19.24 & 139.55 & 18.19& 44.21 & 40.98\\
\midrule 
Ours-LSTM &  \textbf{22.43} & 180.55 & \textbf{22.57}& \textbf{49.36} & 45.61  & 20.84 & 166.78 & 19.81 & \textbf{46.80} & 43.57\\
Ours-transformer & \textbf{22.43} & \textbf{180.66} & 22.54& 49.31 & \textbf{45.66}  & \textbf{20.87} & \textbf{166.82} & \textbf{19.89} & 46.76 & \textbf{43.61} \\
\bottomrule 
\end{tabular}
\label{results:vqa2andcoco}
\end{table*}

\begin{table}[htb]
\centering
\caption{Human Evaluation Results on VQA2.0.}
\begin{tabular}{l|ccc}
\toprule  
Method & Syntax& Semantics& Relevance\\
\midrule  
Radial& 4.25~(0.36) &4.04~(0.45) &3.09~(0.24)\\
Dual& 4.53~(0.40) &4.38~(0.58) & 3.24~(0.26)\\
Ours & \textbf{4.73~(0.35)} &4.61~(0.40) &3.58~(0.39)\\
GT & 4.52~(0.34) &\textbf{4.63~(0.41)} & \textbf{4.14~(0.50)} \\
\bottomrule 
\end{tabular}
\label{results:human}
\end{table}

\begin{table}[htb]
\centering
\caption{Ablation Results on VQA2.0. All accuracies are in \%.}
\begin{tabular}{l|ccc}
\toprule  
Method  &BLEU@4 & METEOR &Q-BLEU@1\\
\midrule 
Full Model(Double hints) & \textbf{22.43} & \textbf{22.57} & \textbf{45.61} \\
\midrule
- visual hints & 21.67 & 21.93 & 44.52\\
\midrule
- visual/answer hints & 18.53 & 20.06 & 40.32\\
  + answer type & \\
  \midrule
- gnn &  22.17 & 22.31 & 45.34\\
\midrule
- gnn + transformer & 22.30 & 22.40 & 45.53 \\
\midrule
- visual attn & 21.98 & 22.26 & 45.11 \\
\midrule
- gnn,visual attn & 21.92 & 22.21 & 45.04\\
\midrule
- pos/ans predict & 22.34 & 22.51 & 45.57\\
\bottomrule 
\end{tabular}
\label{results:ablation}
\end{table}

\section{Experiments}
In this section, we evaluate the effectiveness of our proposed model. The code and data for our model are provided for research purposes\footnotemark[1].
\footnotetext[1]{Please refer to the \href{https://github.com/AlanSwift/DH-VQG}{github repo}.}
\subsection{Datasets and Pre-processing}
\label{sec:dataset}
\textbf{Datasets.} We conduct the experiments on the VQA2.0~\cite{Antol_2015_ICCV} and COCO-QA~\cite{ren2015exploring} datasets. Both of them use images from MS-COCO~\cite{lin2014microsoft}. And to fit our setting, we introduce a simple method to generate the visual hints of the original pairs~(image, question, answer) without human annotations: (1) we use the Mask R-CNN to generate objects~(with category attributes) $\mathcal{R}$ in the image. (2) we use Standford CoreNLP to find the noun-words in both questions and answers. For each object attributes and noun-words we use the GloVe model to initialize them and take the average to get the vector representation denoted by $\mathbf{g}_{obj}$ and $\mathbf{g}_{noun}$.
 The object is recognized as visual hints iff its' l2 distance with any $\mathbf{g}_{noun}$ is smaller than the threshold $\mu$. But there are two special cases that can lead to no aligned objects: (1) there are exactly no visual hints~(eg. Q: Is there any book? A: No) (2) the error caused by the detection model or the NLP tools leads to no visual hints. For the first case, we will keep them. For the second case, we will drop them due to the technical drawback. Moreover, although one image could have multiple answer hints (one image-answer pair generally corresponds to one question), there are a small portion of image-answer pair being linked to multiple questions. In this case, we will randomly reserve one. After pre-processing, the VQA 2.0 contains 239973 examples for training and 116942 for validation, respectively. And the COCO-QA dataset contains 53440 for training and 26405 for validation, respectively.
 
 It is worth noting that the comparison with other baselines under this setting is quite fair since we just use data pre-processing to generate the visual hints without any human annotations. Furthermore, although the generated visual hints are fairly noisy since we simply align them by GloVe l2-distance metric, we will show later that the generated visual hints are still very useful, providing important additional information beyond the corresponding textual answer information.

\noindent \textbf{Pre-processing.}
We employ pre-trained ResNet-101 to extract visual image features. And for each image, we use a Masked-RCNN detector with ResNeXt-101 backbone to detect 100 object regions~(selected by confidence score) and extract features~(fc6). The ResNet-101 is from torchvision and the Mask RCNN is from Detectron2~\cite{wu2019detectron2}. After truncating questions longer than 20 words, we build vocabulary on words with at least 3 occurrences. Since the test split is not open for the public, we divide the train set to 90\% train split and 10\% validation split.

\subsection{Hyper-parameter settings}
\textbf{Overall setting.} The image feature dimension is 2048 and the object feature dimension is 1024. The words' dimension is 512 and their weights are randomly initialized. The hidden size in Eq. 3 is 128 to save CUDA memory. All the hidden size is 1024 if not otherwise specified. As for the visual hints prediction, the $\eta$ is 4 and $\lambda$ is 2. We optimize hyper-parameters with random search~\cite{bergstra2012random}. The overall loss function's $\alpha$ is 0.005, $\beta$ is 0.001 and $\gamma$ is 0.001 for most cases. But for ablation model '-vh', $\alpha$ is 0, $\beta$ is 0.001 and $\gamma$ is 0.002. For ablation model '-position', $\alpha$ is 0.005, $\beta$ is 0 and $\gamma$ is 0.001. And for ablation model '-answer', $\alpha$ is 0.005, $\beta$ is 0.001 and $\gamma$ is 0. When obtaining visual hints, $\mu$ is set 5.7.We adopt adam~\cite{kingma2014adam} optimizer with 0.0002 learning rate. The batch size is 120.

\noindent \textbf{Graph embedding module setting.} The multi-head cosine similarity metric's $k$ is 3. The layer of GCN is 2. The sparsing hyper-parameter $\epsilon$ is 0.75.

\subsection{Baseline Methods and Evaluation Metrics}
\subsubsection{Baseline Methods} We compare against the following baselines in our experiments: 

\paragraph{I2Q} It means generating the questions without any hints. We adopt the classic image caption show attend and tell method~\cite{xu2015show}.

\paragraph{IT2Q} It means generating questions with answer types. We modify the show attend and tell~\cite{xu2015show} method to take input from the joint embedding of the image and answer type. And since there are no answer type annotations in the original datasets, we adopt the same answer-type information as IMVQG which is a baseline we will discuss later.

\paragraph{IMVQG~\cite{krishna2019information}} This is a baseline that maximizes the mutual information among the generated questions, the input images, and the expected answers. They use the answer category~(type) as a hint. For VQA 2.0 dataset, since they just annotate 80\% of the original dataset, so we annotate the rest of them as "other". And for the COCO-QA dataset, we find that there are only 430 answers, so we annotate their type attribute by ourselves just like they do in VQA 2.0.

\paragraph{Dual~\cite{li2018visual}} This is another competitive baseline that views the VQG task as the dual task of VQA based on MUTAN architecture. They train the VQG task along with VQA to enhance both VQG and VQA's performance.

\paragraph{Radial~\cite{xu2020radial}} This is the latest strong baseline for VQG. They use answers to build an answer-radial object graph and learn the graph embedding. Then they use the graph2seq paradigm to generate the questions.

\subsubsection{Evaluation Metrics} Following previous works~\cite{krishna2019information,xu2020radial}, we adopt the standard linguistic measures including BLEU~\cite{papineni2002bleu}, CIDEr~\cite{vedantam2015cider}, METEOR~\cite{banerjee2005meteor}, ROUGE-L~\cite{lin-2004-rouge} and Q-BLEU~\cite{nema2018towards}. 
These scores are calculated by officially released evaluation scripts.

\begin{figure*}[htb]
  \centering
  \includegraphics[width=1\linewidth]{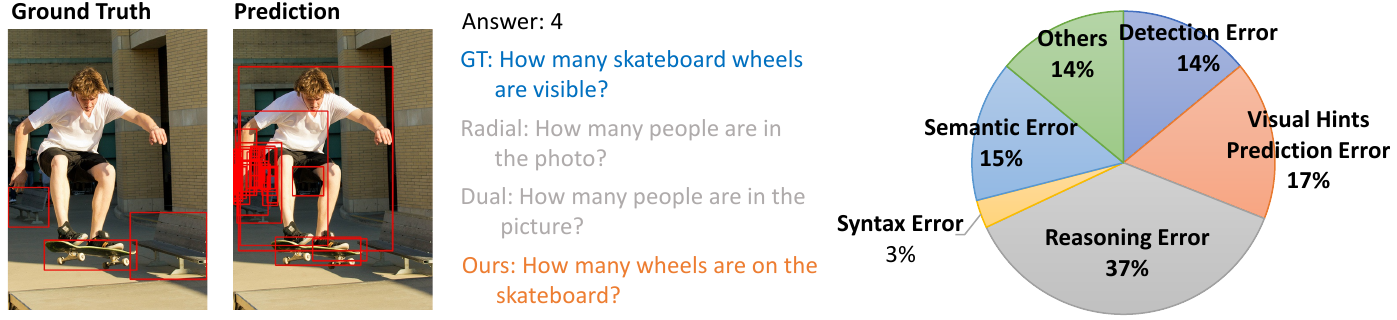}
  \caption{Case study~(left) and error analysis~(right) results.}
  \label{example_and_error}
\end{figure*}

\begin{figure*}[htb]
  \centering
  \includegraphics[width=1\linewidth]{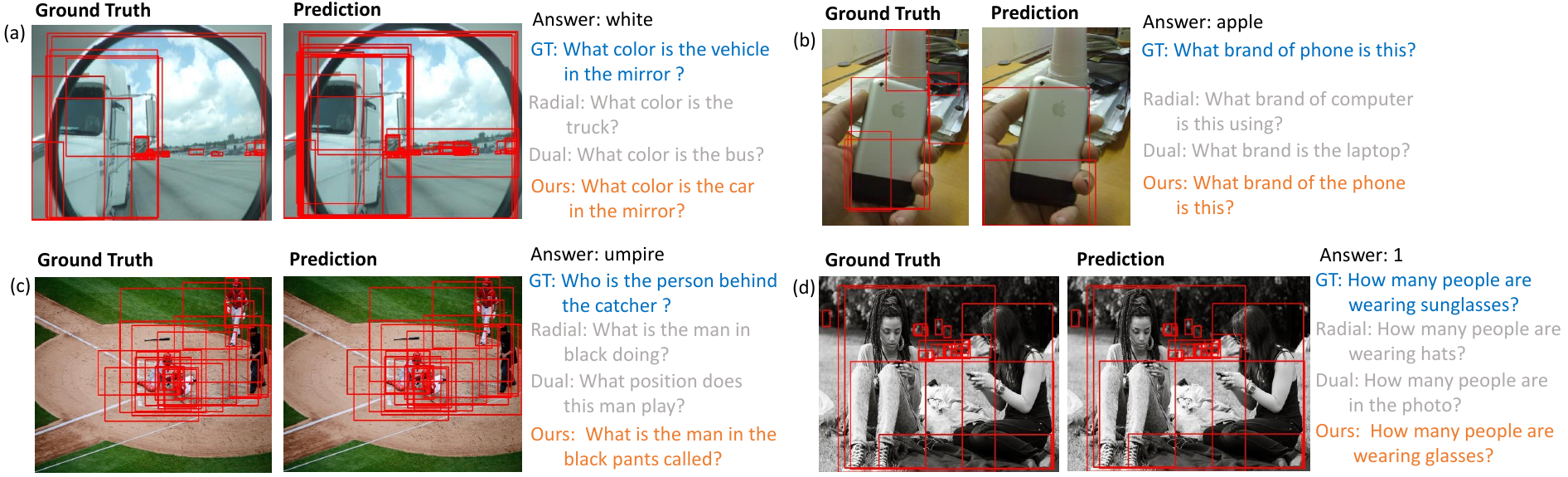}
  \caption{The details of case study examples. The red rectangles mean the visual-hint regions.}
  \label{case_study_full}
\end{figure*}

\subsection{Result Analysis and Human Evaluation}
Table \ref{results:vqa2andcoco} shows the automatic evaluation results comparing our proposed method against other state-of-art baselines. 
We can see that our method consistently outperforms previous methods by a significant margin on both datasets. It highlights that with double hints our proposed method makes a solid step towards addressing the two identified issues described in Sec. \ref{sec:intro}. In addition, we can find that the \myname{} with LSTM decoder (i.e., Ours-LSTM) achieves comparable performance compared with \myname{} with transformer decoder (i.e., Ours-transformer) on both datasets.

Furthermore, we conduct a small-scale human evaluation study on VQA2.0 val set to assess the quality of the questions generated by our method, the ground truth~(GT) and the baselines: Radial and Dual in terms of syntax, semantics, and relevance metrics. 
Concretely, we randomly select 50 examples for each system: 1) the ground-truth results~(denoted as GT), 2) our results~(denoted as Ours), 3) the 'Radial' baseline's results~(denoted as Radial), 4) the 'dual' baseline's results~(denoted as Dual).

We ask 5 human evaluators to give feedback on the quality of questions randomly selected in the results of 4 systems. In each example, given a triple containing a raw image, a target answer, and an anonymized system's output, they are asked to rate the quality of the output by answering the three questions: a) is the question syntactically correct? b) is the question semantically correct? c) is the question relevant to the image and the answer pair? For each question, the rating scale is from 1 to 5. The standard is 1. Pool~(not acceptable), 2. Marginal, 3. Acceptable, 4. Good, 5. Excellent. We develop software to automatically collect the evaluation results. The software will feed the examples and calculate the scores.

The results are shown in Table \ref{results:human}. We report the mean and standard deviation scores. It is clear to see that our model significantly outperforms all strong baseline methods on all metrics, making it the closest one to the ground-truth.

\subsection{Ablation Study}
In this section, we conduct the ablation studies to demonstrate the importance of each key component including double hints, double-hints guided graph, visual attention, and multi-task auto-encoder on the VQA 2.0. Without loss of generality, we use the \myname{} with LSTM decoder as the full model (abbr: Full Model(Double hints)).

Specifically, we will remove one or more components at one time to generate the following ablation models:
(1) w/o. visual hints~(abbr: -visual hints): we remove the visual-hints prediction module.
(2) w/o. double hints~(abbr: -visual/answer hints + answer type): we remove both the visual and answer hints but use the answer-type to further assess the effect of double hints.
(3) w/o. double-hints guided graph~(abbr: -gnn): we remove the implicit graph construction and gcn encoding modules.
(4) replace gnn with transformer encoder~(abbr: - gnn + transformer): we remove the gnn encoder and use the transformer encoder to encode the object features.
(5) w/o. visual attention~(abbr: -visual attention): we remove the non-visual-hint mask during attention when decoding. 
(6) w/o. double-hints guided graph and visual attention~(abbr: gnn, visual attn): We combine the ablation (3) and (4) together to further assess the effect of them.
(7) w/o. position and answer auto-encoder~(abbr: -pos/ans predict): we  set $\beta$ and $\gamma$ to zero for reducing multi-task auto-encoder to visual hints only based auto-encoder.

The ablation study results on VQA2.0 val set's results are shown in Table \ref{results:ablation}.
There are several interesting observations worth noting here. 
Specifically, by turning off the visual hints (-visual hints), the model's performance drops nearly 2.8\% (METEOR). It confirms that the visual hints are indeed helpful for high-quality question generation. In addition, it's interesting to find that it still outperforms the baseline methods. We think the fine-grained interaction of visual objects and answers helps.
And when we further discard the answer hints (-visual/answer hints + answer type), the performance continues to drop rapidly by 18.2\% (BLEU@4), which further demonstrates that the answer hints are a more helpful signal compared to the answer type. These results indicate the importance of double hints.

Next, by turning off the double-hints guided graph (i.e., - gnn), we observe that the performance drops nearly 1.2\% (METEOR). Furthermore, when we replace the gnn with transformer (i.e., - gnn + transformer), the performance still drops nearly 0.8\% (METEOR).
It demonstrates that it is beneficial to exploit the hidden relations by learning a dynamic graph. Furthermore, when we discard the visual attention during decoding (-visual attn), the performance drops by 1.3\% (METEOR). This result illustrates that if we force the model to focus on only the predicted visual-hint object regions, it can generate higher quality questions, which further confirms the impact of visual hints. Additionally, if we turn off both the graph and the visual attention (-gnn,visual attn), the performance continues to drop, which further confirms that both these two components play an indispensable role in the framework. Finally, we observe that if we discard the multi-task auto-encoder (-pos/ans predict), the performance drops slightly. It shows by predicting the object position and target answer, the model can learn more robust visual hint features.

\begin{figure}[htb]
  \centering
  \includegraphics[width=1\linewidth]{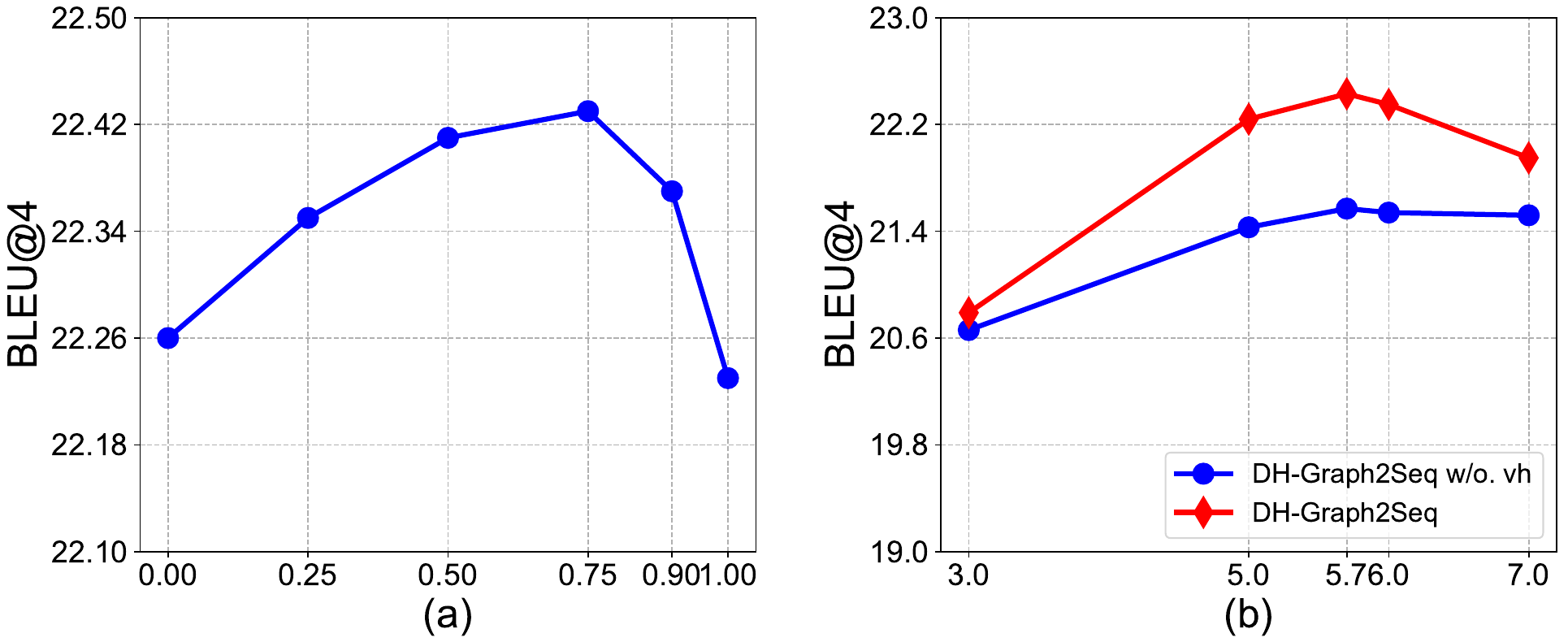}
  \caption{(a) The influence of $\epsilon$ for different graph sparsity and (b) The influence of $\mu$ in visual hints annotating procedure.}
  \label{eps_mu}
\end{figure}

\subsection{The analysis of hyper-parameters}
\subsubsection{The $\epsilon$ analysis}

To study the effect of the $\epsilon$ in Sec. 3.3, we conduct experiments on VQA 2.0 with the $\epsilon$ varying in $[0, 1]$. The results are shown in Figure \ref{eps_mu} (a). We find that the model achieves the best performance when $\epsilon$ ranging in [0.5, 0.8]. It drops rapidly when $\epsilon$ is close to 1 which means the graph is too sparse.
\subsubsection{The $\mu$ analysis}
To study the effect of the $\mu$ in Sec. 4.1, we conduct experiments on VQA 2.0 with the $\mu$ varying in $\{3, 5.7, 7\}$. When $\mu$ is larger, the visual hint will be more accurate, but the reserved questions will be much less. So in order to compare the effect fairly, we use the same test set and evaluate both the \myname{} model and \myname{} w/o visual hint (abbr: \myname{} w/o vh). The results are shown in Figure \ref{eps_mu} (b). We can find that the \myname{} consistently outperforms the \myname{} w/o vh. When $\mu$ is smaller, although the visual hints are more accurate, the processed dataset has less questions which lead to performance drop. And when $\mu$ is larger, although we can reserve more questions, the visual hints' quality is lower. So we think a reasonable choice of $\mu$ is near 5.7.

\subsection{Case Study and Error Analysis}
\label{sec:case_study_and_error_analysis}
\begin{figure}[htb]
  \centering
  \includegraphics[width=0.85\linewidth]{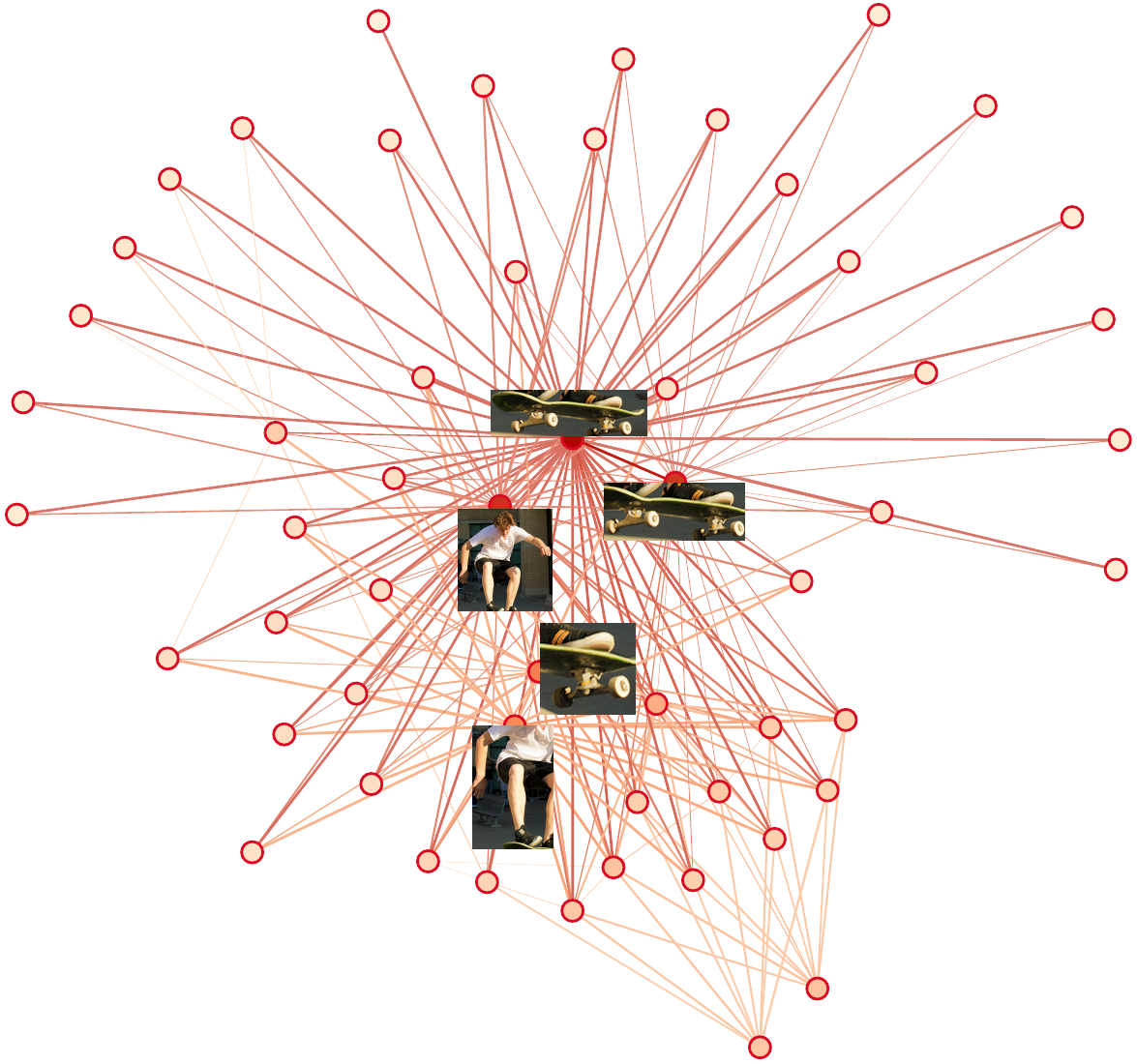}
  \caption{The visualization of the learned graph structure.}
  \label{vis-learnedgraph}
\end{figure}
\begin{figure*}[htb]
  \centering
  \includegraphics[width=1\linewidth]{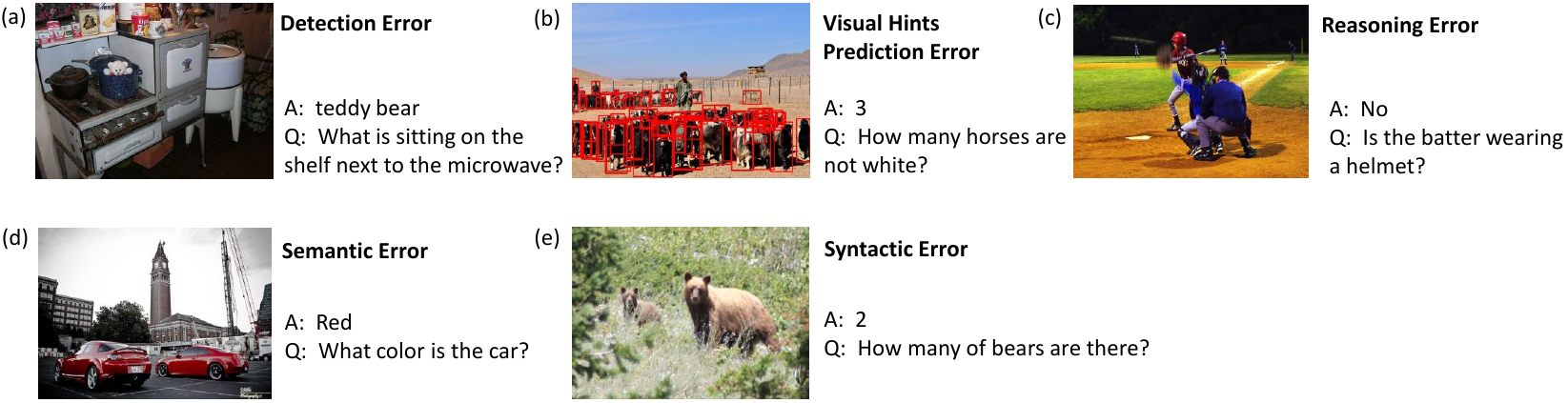}
  \caption{The details of error examples.}
  \label{error_analysis_fig}
\end{figure*}

In this section, we will further perform the case study to illustrate the superiority of our method compared with other baselines. What's more, we also dive into the failures of our model by pinpointing different error cases. The results are shown in Figure \ref{example_and_error}. We further visualize the  learned graph in this example in Figure \ref{vis-learnedgraph}.
By learning the object graph and generating with the guidance of double hints, we can find that our model indeed generates more answer-aware, region-referential, and high-quality questions. We also provide more qualitative cases in Figure \ref{case_study_full}. We can find that our model can generate more complete and vivid questions compared with baseline methods. We think that with the help of double hints, our graph2seq model can find the proper image regions and exploit the rich structure relations better.

For error cases, since it is very difficult to classify a particular example into the certain error type, we are only able to qualitatively divide them into five different error categories. See Figure \ref{error_analysis_fig} for error cases of our results. We present one example of each error reason.

\begin{itemize}
    \item 1) detection error: It means our model recognizes the objects incorrectly. In the picture, the 'teddy bear' is next to the refrigerator but our model recognizes it as 'microwave'.
    \item 2) visual hint prediction error: It means our model predicts the visual hints incorrectly, which largely misleads the generation. Indeed, the answer '3' refers to the number of humans, but the model picks the horses out and overlooks the men.
    \item 3) reasoning error: It means our model infers the relations among the objects incorrectly. In the image, the batter wears the helmet but the expected answer is 'no'.
    \item 4) semantic error: It means our model generates questions which are semantically incorrect. In the image, there are at least two cars, but in the generated question, the car is in the singular form.
    \item 5) syntactic error: It means our model generates questions which are syntactically incorrect. The 'How many of' is incorrect.
\end{itemize}

\begin{figure}[htb]
  \centering
  \includegraphics[width=1\linewidth]{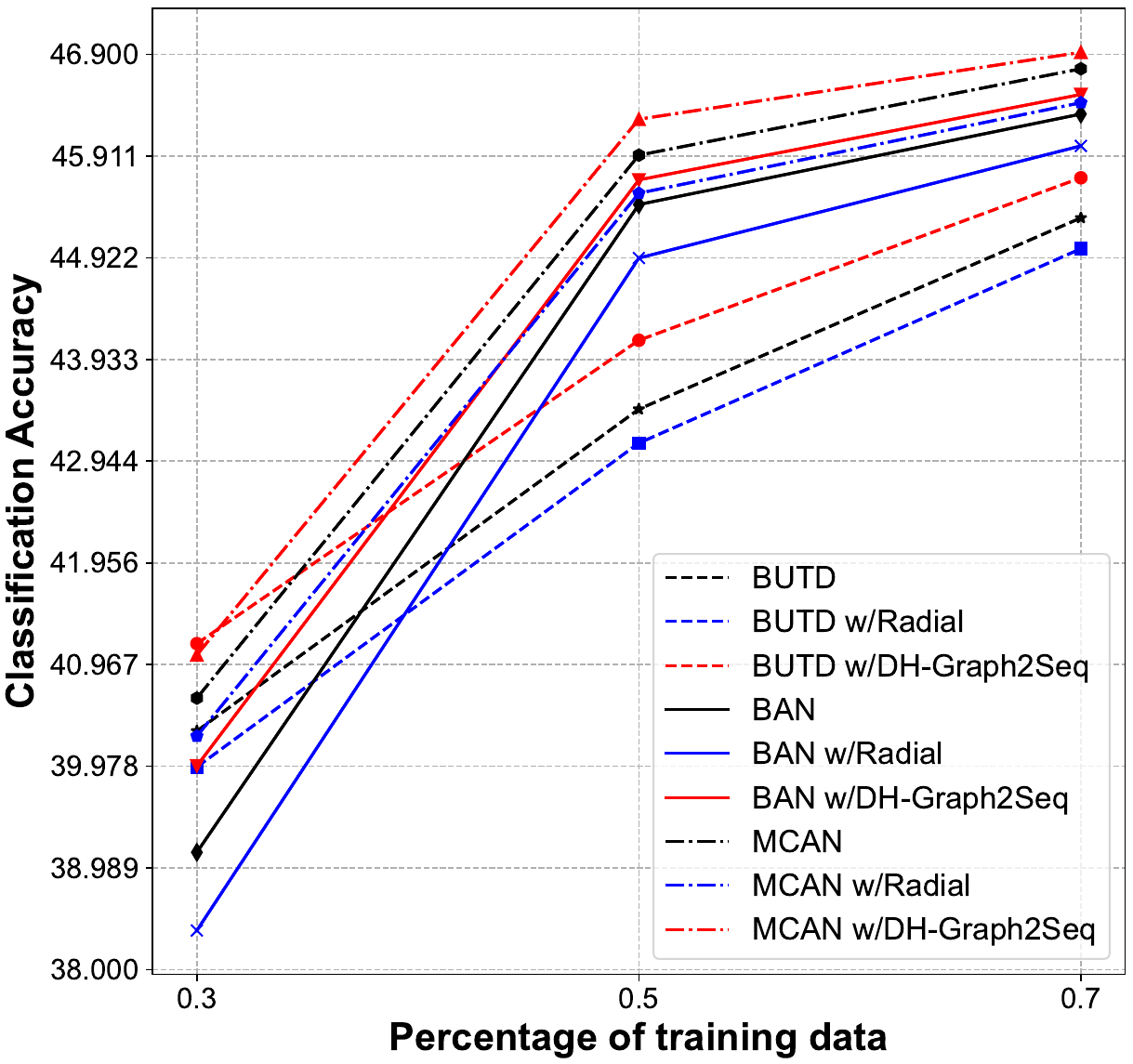}
  \caption{Performance of VQA model when varying the proportions of training data via data augmentation.}
  \label{data_argument}
\end{figure}

\subsection{Data Augmentation}

One of the most important applications of VQG is to provide more training data for VQA. Here, we use the proposed method as data augmentation model to generate more questions for training VQA methods. In particular, we employ three existing strong VQA models: i) Bottom-Up Top-Down~(abbr: BUTD)~\cite{anderson2018bottom}, ii) the Bilinear Attention Network~(abbr: BAN)~\cite{kim2018bilinear} and iii) the Modular Co-Attention Networks (abbr: MCAN)~\cite{Yu_2019_CVPR}. We train these three VQA models on VQA 2.0 dataset and use top-1 accuracy as the evaluation metric. And we choose the best VQG baseline (Radial) and our proposed method as two data augmentation methods. In order to examine the effect of QG-driven data augmentation on the VQA, we compare the performance of the VQA model (i.e. BUTD) with two data augmentation variants, namely, BUTD w/Radial, BUTD w/\myname{}. Specifically, we split the train-set~(the same as the VQG dataset) to $x \in \{0.3, 0.5, 0.7\}$. The VQA models are trained only on the $x$ part, while the other VQA variants are trained on the combination of the golden $x$ part and the questions generated by VQG models. What's more, to train the VQG model, we further split $x$ part to $80\%/20\%$(train/dev). 

As shown in Figure \ref{data_argument}, we observe that both VQG models are able to consistently help improve the VQA performance and the performance boost is the most significant when training data is scarce (i.e., 30\%). Notably, our \myname{} model outperforms the Radial baseline consistently. Surprisingly, we observe that the Radial augmented VQA models perform inferior to the single VQA models slightly. This is because the generated questions may contain noise, which may mislead the VQA model in some ways \cite{xu2020radial}.

\subsection{Zero-shot VQA}
Furthermore, we introduce VQG as an effective solution for zero-shot VQA (ZS-VQA) problems. Here, we employ VQG as a natural way to generate VQA questions, which can be used as valuable data supplementary to address the zero-shot VQA problem. To further evaluate the VQG models under the zero-shot VQA downstream task, we design various combinations from the state-of-art VQA and VQG methods on the VQA2.0 dataset.

Technically, following \cite{xu2020radial,li2018zero}, we firstly obtain the \textit{zero-shot words} the then construct the \textit{zero-shot VQA dataset}. In particular, we mix the original train and test samples together and find the same words between questions and answers by filtering out stop words. Then, after randomly sampling 10\% words from them (denoted as ZS-VQA words), we create the ZS-VQA test data set from all samples with the ZS-VQA words in answer. Then we can obtain the ZS-VQA training set which has no overlap with the ZS-VQA testing set.

After that, we employ two VQG methods: 1) the best baseline Radial and 2) our \myname{} method to produce the questions for the ZS-VQA problem. In particular, we employ two SOTA VQA methods: 1) BUTD and 2) BAN and use top-1 accuracy as the evaluation metric. Specifically, following \cite{xu2020radial}, since ZS-VQA words appear in the question, they can definitely be found in the images, and we treat them as the target answers to generate questions to help the VQA model to predict the zero-shot answer. The results are shown in Table \ref{results:zs-shot}.

From Table \ref{results:zs-shot}, we can observe that both BUTD and BAN VQA methods achieve 0\% accuracy in the ZS-VQA test set. The reason is that during training, the zero-shot answers are not available, thus the models cannot recognize them correctly in the inference. However, when we combine the VQG methods, both the VQA models gain large improvements, which proves that the VQG is potentially helpful for zero-shot VQA. Specifically, we find that the variants with \myname{} consistently outperform those with Radial, which verifies the superiority of our proposed method. 
\begin{table}[t]
\caption{Zero-shot VQA results on zero-shot VQA2.0 test set.\\ All accuracies are in \%.}
\centering
\begin{tabular}{l|c|c|c}
\toprule  
Method & w/o. VQG & Radial & \myname{} \\
\midrule 
BUTD & 0.00 & 18.15 & 18.83\\
BAN & 0.00 &18.37 & 18.91 \\
\bottomrule 
\end{tabular}
\label{results:zs-shot}
\end{table}

\section{Conclusion}
Despite promising results that have been achieved in the visual question generation task, we show in this paper that VQG still suffers from \emph{one-to-many mapping} and \emph{complex implicit relation modelling} problems. To address those issues, we propose a novel setting with double hints for the visual question generation task, which could effectively mitigate the one-to-many mapping issue. Under the new setting, we explicitly cast the VQG task as a Graph-to-Sequence learning problem and design a novel \myname{} model. This model learns an implicit graph topology to capture the rich relationships within an image and then utilizes the Graph2Seq model to generate the questions with answer-awareness and region-reference. Our extensive experiments on VQA2.0 and COCO-QA datasets demonstrate that our proposed model can significantly outperform existing state-of-the-art by a large margin. Furthermore, the data augmentation and zero-shot VQA experiments demonstrate that the \myname{} can effectively help real-world applications. We hope that our work will encourage further exploration of VQG to generate more vivid and meaningful questions.

\section{Limitation and Future Works}

Despite the significant progress achieved by \myname{}, several challenges and areas for improvement remain:
\begin{itemize}
    \item \textbf{Visual Hint Quality.} The quality of visual hints plays a pivotal role in question generation quality. While we have introduced a rule-based method for generating visual hints without requiring human annotation, this approach is susceptible to noise and inaccuracies, which can adversely affect generation performance. In future work, we aim to explore more effective methods to enhance the quality of visual hints, thereby improving overall question generation accuracy.
    \item \textbf{Visual-Language Pretraining.} Although \myname{} has advanced the quality of visual question generation with the double hints guided learning paradigm and graph-to-sequence framework, challenges persist in visual-language reasoning. We think that current limitations in data and model scale may impede effective multi-modal reasoning. In the future, we will extend our approach through large-scale pretraining, involving the collection of more extensive datasets to bolster the quality of question generation and enhance visual-language reasoning capabilities.
\end{itemize}


%



\ifCLASSOPTIONcompsoc
  \section*{Acknowledgments}
\else
  \section*{Acknowledgment}
\fi

This work was supported by the Key Research and Development Projects in Zhejiang Province (No. 2024C01106), the NSFC (No. 62272411), the National Key Research and Development Project of China (2018AAA0101900),  the Tencent WeChat Rhino-Bird Special Research Program (Tencent WXG-FR-2023-10), and Research funding from FinVolution Group.

\ifCLASSOPTIONcaptionsoff
  \newpage
\fi



\bibliographystyle{IEEEtran}
\bibliography{main}

\begin{thebibliography}{10}
\providecommand{\url}[1]{#1}
\csname url@samestyle\endcsname
\providecommand{\newblock}{\relax}
\providecommand{\bibinfo}[2]{#2}
\providecommand{\BIBentrySTDinterwordspacing}{\spaceskip=0pt\relax}
\providecommand{\BIBentryALTinterwordstretchfactor}{4}
\providecommand{\BIBentryALTinterwordspacing}{\spaceskip=\fontdimen2\font plus
\BIBentryALTinterwordstretchfactor\fontdimen3\font minus \fontdimen4\font\relax}
\providecommand{\BIBforeignlanguage}[2]{{%
\expandafter\ifx\csname l@#1\endcsname\relax
\typeout{** WARNING: IEEEtran.bst: No hyphenation pattern has been}%
\typeout{** loaded for the language `#1'. Using the pattern for}%
\typeout{** the default language instead.}%
\else
\language=\csname l@#1\endcsname
\fi
#2}}
\providecommand{\BIBdecl}{\relax}
\BIBdecl

\bibitem{li2018visual}
Y.~Li, N.~Duan, B.~Zhou, X.~Chu, W.~Ouyang, X.~Wang, and M.~Zhou, ``Visual question generation as dual task of visual question answering,'' in \emph{Proceedings of the IEEE Conference on Computer Vision and Pattern Recognition}, 2018, pp. 6116--6124.

\bibitem{10.1109/TPAMI.2017.2754246}
\BIBentryALTinterwordspacing
P.~Wang, Q.~Wu, C.~Shen, A.~Dick, and A.~van~den Hengel, ``Fvqa: Fact-based visual question answering,'' \emph{IEEE Trans. Pattern Anal. Mach. Intell.}, vol.~40, no.~10, p. 2413–2427, Oct. 2018. [Online]. Available: \url{https://doi.org/10.1109/TPAMI.2017.2754246}
\BIBentrySTDinterwordspacing

\bibitem{7934440}
Q.~Wu, C.~Shen, P.~Wang, A.~Dick, and A.~v.~d. Hengel, ``Image captioning and visual question answering based on attributes and external knowledge,'' \emph{IEEE Transactions on Pattern Analysis and Machine Intelligence}, vol.~40, no.~6, pp. 1367--1381, 2018.

\bibitem{Jain_2018_CVPR}
U.~Jain, S.~Lazebnik, and A.~G. Schwing, ``Two can play this game: Visual dialog with discriminative question generation and answering,'' in \emph{Proceedings of the IEEE Conference on Computer Vision and Pattern Recognition (CVPR)}, June 2018.

\bibitem{mora2016towards}
I.~M. Mora, S.~P. de~la Puente, and X.~G.-i. Nieto, ``Towards automatic generation of question answer pairs from images,'' \emph{CVPRW}, 2016.

\bibitem{krishna2019information}
R.~Krishna, M.~Bernstein, and L.~Fei-Fei, ``Information maximizing visual question generation,'' in \emph{Proceedings of the IEEE Conference on Computer Vision and Pattern Recognition}, 2019, pp. 2008--2018.

\bibitem{mostafazadeh2016generating}
\BIBentryALTinterwordspacing
N.~Mostafazadeh, I.~Misra, J.~Devlin, M.~Mitchell, X.~He, and L.~Vanderwende, ``Generating natural questions about an image,'' in \emph{Proceedings of the 54th Annual Meeting of the Association for Computational Linguistics (Volume 1: Long Papers)}.\hskip 1em plus 0.5em minus 0.4em\relax Berlin, Germany: Association for Computational Linguistics, Aug. 2016, pp. 1802--1813. [Online]. Available: \url{https://aclanthology.org/P16-1170}
\BIBentrySTDinterwordspacing

\bibitem{9126124}
L.~Peng, Y.~Yang, Z.~Wang, Z.~Huang, and H.~T. Shen, ``Mra-net: Improving vqa via multi-modal relation attention network,'' \emph{IEEE Transactions on Pattern Analysis and Machine Intelligence}, pp. 1--1, 2020.

\bibitem{8847465}
Q.~Cao, X.~Liang, B.~Li, and L.~Lin, ``Interpretable visual question answering by reasoning on dependency trees,'' \emph{IEEE Transactions on Pattern Analysis and Machine Intelligence}, vol.~43, no.~3, pp. 887--901, 2021.

\bibitem{8528867}
F.~Liu, T.~Xiang, T.~M. Hospedales, W.~Yang, and C.~Sun, ``Inverse visual question answering: A new benchmark and vqa diagnosis tool,'' \emph{IEEE Transactions on Pattern Analysis and Machine Intelligence}, vol.~42, no.~2, pp. 460--474, 2020.

\bibitem{jain2017creativity}
U.~Jain, Z.~Zhang, and A.~G. Schwing, ``Creativity: Generating diverse questions using variational autoencoders,'' in \emph{Proceedings of the IEEE Conference on Computer Vision and Pattern Recognition}, 2017, pp. 6485--6494.

\bibitem{zhang2016automatic}
\BIBentryALTinterwordspacing
S.~Zhang, L.~Qu, S.~You, Z.~Yang, and J.~Zhang, ``Automatic generation of grounded visual questions,'' in \emph{Proceedings of the Twenty-Sixth International Joint Conference on Artificial Intelligence, {IJCAI-17}}, 2017, pp. 4235--4243. [Online]. Available: \url{https://doi.org/10.24963/ijcai.2017/592}
\BIBentrySTDinterwordspacing

\bibitem{liu2018ivqa}
F.~Liu, T.~Xiang, T.~M. Hospedales, W.~Yang, and C.~Sun, ``ivqa: Inverse visual question answering,'' in \emph{Proceedings of the IEEE Conference on Computer Vision and Pattern Recognition}, 2018, pp. 8611--8619.

\bibitem{shah2019cycle}
M.~Shah, X.~Chen, M.~Rohrbach, and D.~Parikh, ``Cycle-consistency for robust visual question answering,'' in \emph{Proceedings of the IEEE Conference on Computer Vision and Pattern Recognition}, 2019, pp. 6649--6658.

\bibitem{xu2020radial}
X.~Xu, T.~Wang, Y.~Yang, A.~Hanjalic, and H.~T. Shen, ``Radial graph convolutional network for visual question generation,'' \emph{IEEE Transactions on Neural Networks and Learning Systems}, 2020.

\bibitem{kipf2016semi}
\BIBentryALTinterwordspacing
T.~N. Kipf and M.~Welling, ``{Semi-Supervised Classification with Graph Convolutional Networks},'' in \emph{Proceedings of the 5th International Conference on Learning Representations}, ser. ICLR '17, 2017. [Online]. Available: \url{https://openreview.net/forum?id=SJU4ayYgl}
\BIBentrySTDinterwordspacing

\bibitem{gilmer2017neural}
J.~Gilmer, S.~S. Schoenholz, P.~F. Riley, O.~Vinyals, and G.~E. Dahl, ``Neural message passing for quantum chemistry,'' in \emph{Proceedings of the 34th International Conference on Machine Learning-Volume 70}.\hskip 1em plus 0.5em minus 0.4em\relax JMLR. org, 2017, pp. 1263--1272.

\bibitem{xu2018graph2seq}
K.~Xu, L.~Wu, Z.~Wang, Y.~Feng, M.~Witbrock, and V.~Sheinin, ``Graph2seq: Graph to sequence learning with attention-based neural networks,'' \emph{arXiv preprint arXiv:1804.00823}, 2018.

\bibitem{chen2019reinforcement}
\BIBentryALTinterwordspacing
Y.~Chen, L.~Wu, and M.~J. Zaki, ``Reinforcement learning based graph-to-sequence model for natural question generation,'' in \emph{International Conference on Learning Representations}, 2020. [Online]. Available: \url{https://openreview.net/forum?id=HygnDhEtvr}
\BIBentrySTDinterwordspacing

\bibitem{gao2019dyngraph2seq}
Y.~Gao, L.~Wu, H.~Homayoun, and L.~Zhao, ``Dyngraph2seq: Dynamic-graph-to-sequence interpretable learning for health stage prediction in online health forums,'' \emph{arXiv preprint arXiv:1908.08497}, 2019.

\bibitem{norcliffe2018learning}
W.~Norcliffe-Brown, S.~Vafeias, and S.~Parisot, ``Learning conditioned graph structures for interpretable visual question answering,'' in \emph{Advances in Neural Information Processing Systems}, 2018, pp. 8334--8343.

\bibitem{chen2019graphflow}
\BIBentryALTinterwordspacing
Y.~Chen, L.~Wu, and M.~J. Zaki, ``Graphflow: Exploiting conversation flow with graph neural networks for conversational machine comprehension,'' in \emph{Proceedings of the Twenty-Ninth International Joint Conference on Artificial Intelligence, {IJCAI-20}}, C.~Bessiere, Ed.\hskip 1em plus 0.5em minus 0.4em\relax International Joint Conferences on Artificial Intelligence Organization, 7 2020, pp. 1230--1236, main track. [Online]. Available: \url{https://doi.org/10.24963/ijcai.2020/171}
\BIBentrySTDinterwordspacing

\bibitem{he2016deep}
K.~He, X.~Zhang, S.~Ren, and J.~Sun, ``Deep residual learning for image recognition,'' in \emph{Proceedings of the IEEE conference on computer vision and pattern recognition}, 2016, pp. 770--778.

\bibitem{anderson2018bottom}
P.~Anderson, X.~He, C.~Buehler, D.~Teney, M.~Johnson, S.~Gould, and L.~Zhang, ``Bottom-up and top-down attention for image captioning and visual question answering,'' in \emph{Proceedings of the IEEE conference on computer vision and pattern recognition}, 2018, pp. 6077--6086.

\bibitem{lin2017focal}
T.-Y. Lin, P.~Goyal, R.~Girshick, K.~He, and P.~Doll{\'a}r, ``Focal loss for dense object detection,'' in \emph{Proceedings of the IEEE international conference on computer vision}, 2017, pp. 2980--2988.

\bibitem{chen2019deep}
\BIBentryALTinterwordspacing
Y.~Chen, L.~Wu, and M.~Zaki, ``Iterative deep graph learning for graph neural networks: Better and robust node embeddings,'' in \emph{Advances in Neural Information Processing Systems}, H.~Larochelle, M.~Ranzato, R.~Hadsell, M.~F. Balcan, and H.~Lin, Eds., vol.~33.\hskip 1em plus 0.5em minus 0.4em\relax Curran Associates, Inc., 2020, pp. 19\,314--19\,326. [Online]. Available: \url{https://proceedings.neurips.cc/paper/2020/file/e05c7ba4e087beea9410929698dc41a6-Paper.pdf}
\BIBentrySTDinterwordspacing

\bibitem{lu2018neural}
J.~Lu, J.~Yang, D.~Batra, and D.~Parikh, ``Neural baby talk,'' in \emph{Proceedings of the IEEE conference on computer vision and pattern recognition}, 2018, pp. 7219--7228.

\bibitem{bahdanau2014neural}
D.~Bahdanau, K.~Cho, and Y.~Bengio, ``Neural machine translation by jointly learning to align and translate,'' \emph{arXiv preprint arXiv:1409.0473}, 2014.

\bibitem{Antol_2015_ICCV}
S.~Antol, A.~Agrawal, J.~Lu, M.~Mitchell, D.~Batra, C.~L. Zitnick, and D.~Parikh, ``Vqa: Visual question answering,'' in \emph{The IEEE International Conference on Computer Vision (ICCV)}, December 2015.

\bibitem{ren2015exploring}
M.~Ren, R.~Kiros, and R.~Zemel, ``Exploring models and data for image question answering,'' in \emph{Advances in neural information processing systems}, 2015, pp. 2953--2961.

\bibitem{lin2014microsoft}
T.-Y. Lin, M.~Maire, S.~Belongie, J.~Hays, P.~Perona, D.~Ramanan, P.~Doll{\'a}r, and C.~L. Zitnick, ``Microsoft coco: Common objects in context,'' in \emph{European conference on computer vision}.\hskip 1em plus 0.5em minus 0.4em\relax Springer, 2014, pp. 740--755.

\bibitem{wu2019detectron2}
Y.~Wu, A.~Kirillov, F.~Massa, W.-Y. Lo, and R.~Girshick, ``Detectron2,'' \url{https://github.com/facebookresearch/detectron2}, 2019.

\bibitem{bergstra2012random}
J.~Bergstra and Y.~Bengio, ``Random search for hyper-parameter optimization.'' \emph{Journal of machine learning research}, vol.~13, no.~2, 2012.

\bibitem{kingma2014adam}
D.~P. Kingma and J.~Ba, ``Adam: A method for stochastic optimization,'' \emph{arXiv preprint arXiv:1412.6980}, 2014.

\bibitem{xu2015show}
K.~Xu, J.~Ba, R.~Kiros, K.~Cho, A.~Courville, R.~Salakhudinov, R.~Zemel, and Y.~Bengio, ``Show, attend and tell: Neural image caption generation with visual attention,'' in \emph{International conference on machine learning}, 2015, pp. 2048--2057.

\bibitem{papineni2002bleu}
K.~Papineni, S.~Roukos, T.~Ward, and W.-J. Zhu, ``Bleu: a method for automatic evaluation of machine translation,'' in \emph{Proceedings of the 40th annual meeting on association for computational linguistics}.\hskip 1em plus 0.5em minus 0.4em\relax Association for Computational Linguistics, 2002, pp. 311--318.

\bibitem{vedantam2015cider}
R.~Vedantam, C.~Lawrence~Zitnick, and D.~Parikh, ``Cider: Consensus-based image description evaluation,'' in \emph{Proceedings of the IEEE conference on computer vision and pattern recognition}, 2015, pp. 4566--4575.

\bibitem{banerjee2005meteor}
S.~Banerjee and A.~Lavie, ``Meteor: An automatic metric for mt evaluation with improved correlation with human judgments,'' in \emph{Proceedings of the acl workshop on intrinsic and extrinsic evaluation measures for machine translation and/or summarization}, 2005, pp. 65--72.

\bibitem{lin-2004-rouge}
\BIBentryALTinterwordspacing
C.-Y. Lin, ``{ROUGE}: A package for automatic evaluation of summaries,'' in \emph{Text Summarization Branches Out}.\hskip 1em plus 0.5em minus 0.4em\relax Barcelona, Spain: Association for Computational Linguistics, Jul. 2004, pp. 74--81. [Online]. Available: \url{https://www.aclweb.org/anthology/W04-1013}
\BIBentrySTDinterwordspacing

\bibitem{nema2018towards}
\BIBentryALTinterwordspacing
P.~Nema and M.~M. Khapra, ``Towards a better metric for evaluating question generation systems,'' in \emph{Proceedings of the 2018 Conference on Empirical Methods in Natural Language Processing}.\hskip 1em plus 0.5em minus 0.4em\relax Brussels, Belgium: Association for Computational Linguistics, Oct.-Nov. 2018, pp. 3950--3959. [Online]. Available: \url{https://aclanthology.org/D18-1429}
\BIBentrySTDinterwordspacing

\bibitem{kim2018bilinear}
J.-H. Kim, J.~Jun, and B.-T. Zhang, ``Bilinear attention networks,'' \emph{arXiv preprint arXiv:1805.07932}, 2018.

\bibitem{Yu_2019_CVPR}
Z.~Yu, J.~Yu, Y.~Cui, D.~Tao, and Q.~Tian, ``Deep modular co-attention networks for visual question answering,'' in \emph{Proceedings of the IEEE/CVF Conference on Computer Vision and Pattern Recognition (CVPR)}, June 2019.

\bibitem{li2018zero}
Y.~Li, Y.~Yang, J.~Wang, and W.~Xu, ``Zero-shot transfer vqa dataset,'' \emph{arXiv preprint arXiv:1811.00692}, 2018.

\end{thebibliography}

\end{document}